\documentclass[10pt,twocolumn,letterpaper]{article}

\usepackage[pagenumbers]{arxivstyle}

\usepackage{colortbl}
\usepackage{booktabs,multirow}
\usepackage{makecell}
\usepackage{tabulary}
\usepackage{amsmath}
\usepackage{fontawesome5}
\usepackage{bbding}

\newlength\savewidth\newcommand\shline{\noalign{\global\savewidth\arrayrulewidth
  \global\arrayrulewidth 1pt}\hline\noalign{\global\arrayrulewidth\savewidth}}

\newcommand{\expec}{\mathbb{E}}
\newcommand{\encoder}{\mathcal{E}}
\newcommand{\model}{\epsilon_\theta}
\usepackage{capt-of}
\usepackage[dvipsnames]{xcolor}
\usepackage{xspace}
\usepackage{enumitem}
\usepackage{amsmath,amssymb,amsbsy,amsfonts,dsfont,pifont,bm,bbm,mathrsfs,mathtools,nicefrac}
\usepackage{algorithm,algpseudocode,listings}
\usepackage{booktabs,multirow,adjustbox,diagbox,threeparttable,tabularray}

\newcommand{\method}{\texttt{MotionStone}\xspace}

\newcommand{\tocite}[1]{{\color{red} [TO CITE]}}

\definecolor{CQColor}{rgb}{0.0,0.0,1.0} %

\definecolor{CQRColor}{rgb}{1.0,0.0,1.0} %

\definecolor{gongbiaoblue}{rgb}{0.21,0.49,0.74}
\usepackage[pagebackref,breaklinks,colorlinks,allcolors=gongbiaoblue]{hyperref}
\usepackage{wrapfig}
\usepackage{xspace}
\usepackage[capitalize]{cleveref}  %
\crefname{section}{Sec.}{Secs.}
\Crefname{section}{Section}{Sections}
\crefname{table}{Tab.}{Tabs.}
\Crefname{table}{Table}{Tables}
\crefname{figure}{Fig.}{Figs.}
\Crefname{figure}{Figure}{Figures}
\crefname{equation}{Eq.}{Eqs.}
\Crefname{equation}{Equation}{Equations}
\renewcommand{\thefootnote}{\fnsymbol{footnote}} %

\title{MotionStone: Decoupled Motion Intensity Modulation with Diffusion Transformer for Image-to-Video Generation}

\author{Shuwei Shi$^{1}$\footnotemark[1]\>\,, Biao Gong$^{2}$\footnotemark[2]\>\,, Xi Chen$^{3}$, Dandan Zheng$^{2}$, Shuai Tan$^{2}$, Zizheng Yang$^{2}$,\\ Yuyuan Li$^{4}$, Jingwen He$^{5}$, Kecheng Zheng$^{2}$, Jingdong Chen$^{2}$, Ming Yang$^{2}$, Yinqiang Zheng$^{1}$\footnotemark[3]\\[3pt]
{\normalsize$^1$The University of Tokyo\ \ $^2$Ant Group\ \ $^3$Tongyi Lab}\\[-3pt]
{\normalsize$^4$Zhejiang University\ \ $^5$The Chinese University of Hong Kong}\\
}

\begin{document}

\twocolumn[{
\maketitle
\begin{center}
        \vspace{-8mm}
\includegraphics[width=1\linewidth]{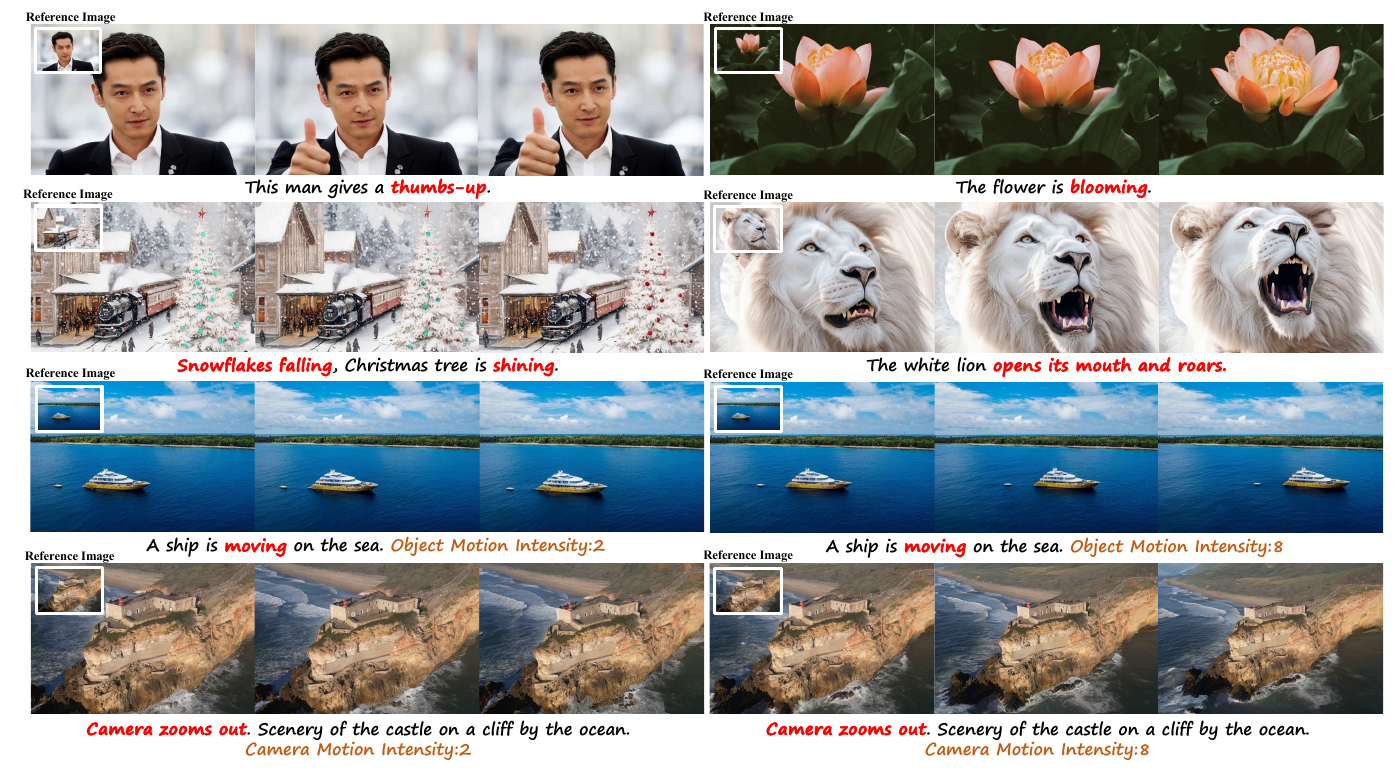}
        \vspace{-7mm}
    \captionof{figure}{
        Samples generated by \method. Our model achieves accurate motion instruction following (\textit{rows-1} and \textit{rows-2}), and is controllable, easily adapting to specified object motion intensities (\textit{row-3}) and camera motion intensities (\textit{row-4}).
    }
    \label{fig:teaser}
\end{center}
}]
\footnotetext[1]{Work done during internship at Ant Group. \footnotemark[2]Project lead.}
\footnotetext[3]{Corresponding author.}

\begin{abstract}

The image-to-video (I2V) generation is conditioned on the static image, which has been enhanced recently by the motion intensity as an additional control signal. These motion-aware models are appealing to generate diverse motion patterns, yet there lacks a reliable motion estimator for training such models on large-scale video set in the wild. Traditional metrics, \textit{e.g.}, SSIM or optical flow, are hard to generalize to arbitrary videos, while, it is very tough for human annotators to label the abstract motion intensity neither. Furthermore, the motion intensity shall reveal both local object motion and global camera movement, which has not been studied before.
This paper addresses the challenge with a new motion estimator, capable of measuring the decoupled motion intensities of objects and cameras in video.
We leverage the contrastive learning on randomly paired videos and distinguish the video with greater motion intensity.
Such a paradigm is friendly for annotation and easy to scale up to achieve stable performance on motion estimation.
We then present a new I2V model, named \method, developed with the decoupled motion estimator.
Experimental results demonstrate the stability of the proposed motion estimator and the state-of-the-art performance of \method on I2V generation. 
These advantages warrant the decoupled motion estimator to serve as a general plug-in enhancer for both data processing and video generation training.

\end{abstract}
\vspace{-10mm}
    
\section{Introduction}
\label{sec:intro}

Image-to-Video (I2V) generation animates static images into fun creative videos which has attracted broad interests in research and industry~\cite{guo2023animatediff, hu2024animate, wang2022latent, zhang2024pia, ren2024consisti2v, xing2025dynamicrafter}.
The key to achieving high-quality I2V results lies in synthesizing sufficient temporal dynamics, which requires effective frame-to-frame motion modeling. Some methods~\cite{chai2023stablevideo, wang2024videocomposer, esser2023structure, wu2023tune, liew2023magicedit, mou2024revideo, zhang2024tora, niu2024mofa} introduce additional conditions into diffusion models, \textit{e.g.,} optical flow, motion trajectories, or depth maps, to better capture motion dynamics.
However, these methods require complex and hard-to-obtain control conditions as inputs, and the training data must be meticulously preprocessed for model training, preventing them from reliably generalizing to videos in the wild.

Recently, several I2V works~\cite{dai2023animateanything, chen2025livephoto, ma2024cinemo} explore text-based motion control and introduce motion intensity as the essential control signal on motion patterns. For example, LivePhoto~\cite{chen2025livephoto} and Cinemo~\cite{ma2024cinemo} leverage text prompts to direct motion and integrate SSIM~\cite{wang2004image} to modulate motion intensity. 
Although these motion-aware models demonstrate improved controllability in motion and enhanced generation quality, the estimation of motion intensity remains inadequate due to the discrepancy between their motion modeling strategy and human motion perception. As a result, the diffusion model is unable to accurately capture the real motion intensity in a video clip during training, which negatively impacts the convergence process.

\begin{figure}[!t]
  \centering
    \includegraphics[width=1\linewidth]{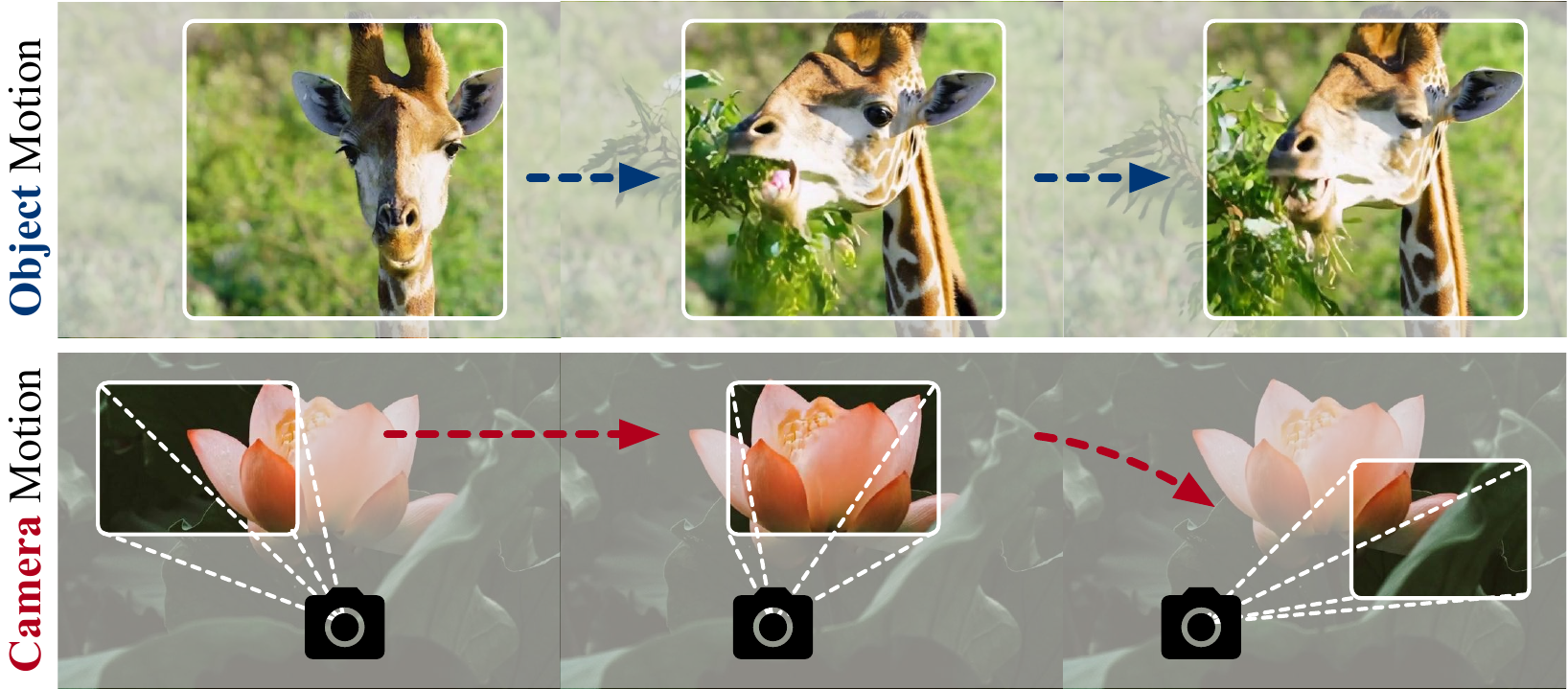}
    \caption{\textbf{Illustration of the motion decoupling}. Decoupling these two types of motion helps the diffusion model learn specific motion patterns, thereby improving the dynamics and controllability of the generated video.} %
    \label{fig:motivation}
    \vspace{-4mm}
\end{figure}

Furthermore, as shown in Fig.~\ref{fig:motivation}, motion patterns in real-world videos could be very complicated including both object motion and camera movement. 
Applying traditional motion extractors, which are not specifically designed for video motion modeling, to estimate motion across entire videos, is unattainable to distinguish between different types of motion, thereby limiting precise control over motion dynamics.
A straightforward way is to learn a motion estimator to predict human perception of object and camera motion intensity in videos.

In this paper, we introduce \method, a general I2V diffusion model to enable decoupled modeling and control of video motion. 
The core of \method is the independent motion estimator, comprising a motion modeling backbone and dual heads to disentangle object and camera motion.  
Specifically, we first propose a video motion annotation method, which requires human annotators to distinguish the relative motion intensity of objects and cameras in randomly selected video pairs. 
Then the proposed motion estimator is trained using a contrastive learning strategy with these relatively annotated video pairs.
For the structure of the motion estimator, we employ a learnable TAdaConv~\cite{huang2021tada} as the motion feature extractor, integrating the pairwise ranking loss and MLP-based motion heads to facilitate motion disentanglement. 

During the training phase of the diffusion model, we freeze the pre-trained motion estimator and use its prediction result as an additional input for noise prediction at each step. In particular, we design a decoupled motion score injection method that allows the model to discern whether each motion intensity control signal originates from the camera or the object, thus achieving decoupled motion modeling in training.
Extensive quantitative and qualitative results demonstrate that \method achieves state-of-the-art performance in text-guided motion control through its decoupled motion intensity guidance and conditional injection, as shown in Fig.~\ref{fig:teaser}. 
\method animates diverse real-world images across various domains, skillfully decomposing motion into object and camera components. 
With its decoupled motion guidance, \method allows users to customize motion intensity, enabling a wide range of motion effects.

\section{Related Work}
\label{sec:related}

\begin{figure*}[!t]
  \centering
    \includegraphics[width=1\linewidth]{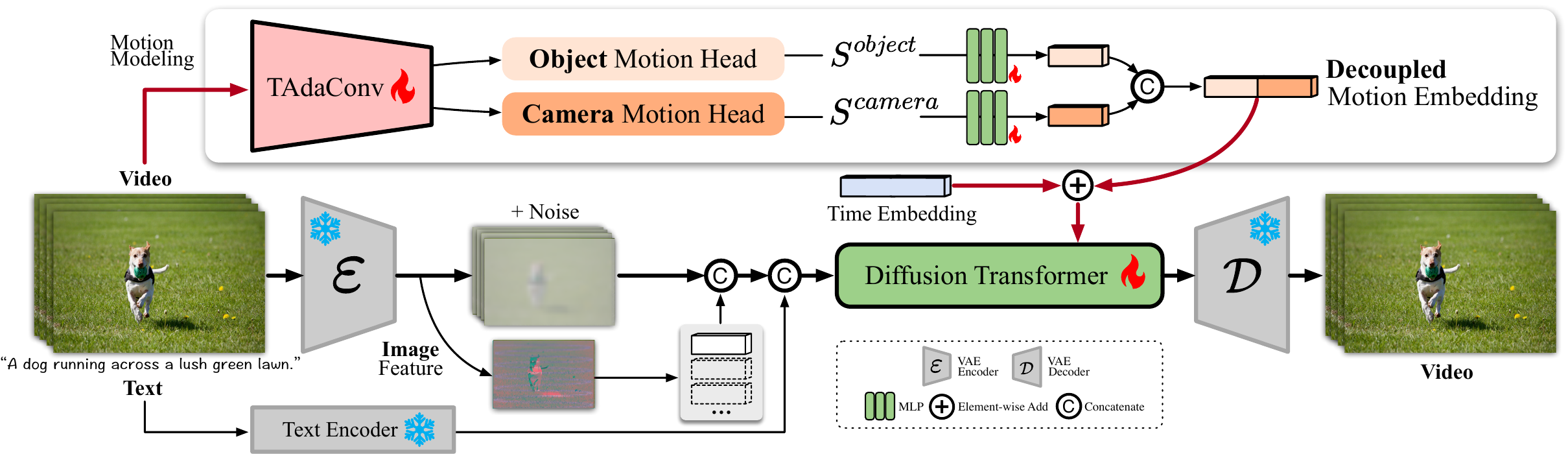}
    \caption{\textbf{The framework of \method}. The first frame of the video serves as the conditioning image, while object and camera motion intensities (ranging from $1$ to $10$) are predicted by the motion estimator and can be customized by users during inference. At the top, the object and camera motion intensities predicted by the motion estimator are processed through an MLP respectively to obtain corresponding embeddings, which are then concatenated along the channel dimension to form the Decoupled Motion Embedding. This embedding is added to the time embedding and injected into the Diffusion Transformer to generate videos.
    }
    \label{fig:framework}
    \vspace{-3mm}
\end{figure*}

\noindent\textbf{Image Animation.} 
Image animation aims to generate controllable videos using a static image as content conditioning. Early methods~\cite{siarohin2021motion, chuang2005animating} focus on modeling motion patterns for specific object types, limiting their ability to generalize motion control to other scenarios. To capture realistic motion from real videos, some methods~\cite{cheng2020time, shalev2022image, siarohin2019first, zhao2022thin, zhao2021sparse,wang2022latent} use a set of videos with various motion patterns as references, transferring these motion patterns to images within the same category. Other approaches model motion for specific scenes, such as fluids~\cite{mahapatra2022controllable,okabe2009animating} and human hair~\cite{xiao2023automatic}. 
Additionally, some methods~\cite{hu2024animate, chang2023magicpose, wang2024unianimate, tan2024animate} convert the human pose into additional conditions, such as depth maps or skeletal points, to guide video generation. 
Although these methods achieve continuous motion control within a specific domain, their applicability remains limited due to the restricted training data and the frequent need for side control signals. Subsequently, some generalizable I2V models~\cite{zhang2023i2vgen, blattmann2023stable} typically train on video data based on pre-trained I2V models. However, the generated videos often exhibit limited motion diversity due to structural limitations and conditions~\cite{chen2025livephoto}.

Recent methods~\cite{zhang2024pia, chen2025livephoto, dai2023animateanything, ma2024cinemo} explore using text as a condition to control motion in video. For instance, PIA~\cite{zhang2024pia} attempts to animate specific domain images using text descriptions of motion. Other methods~\cite{chen2025livephoto, dai2023animateanything, ma2024cinemo} further incorporate coarse-grained motion intensity estimates to generate videos with varying intensities or speeds. However, these approaches lack alignment with human perception, leading to suboptimal results in motion control. In this work, we propose a generalizable framework that uses flexible text as a guiding condition, enabling precise motion modeling in generated videos.

\noindent\textbf{Text-to-Video Generation.} 
The text-to-video (T2V) models have made significant progress along the emerging diffusion models~\cite{ho2020denoising, song2020denoising, gong2024check, shi2024resmaster}. Early T2V models~\cite{wu2023tune, guo2023animatediff, blattmann2023align, wang2023lavie, ma2024latte} harness the strong priors of existing text-to-image (T2I) models, adapting temporal modules trained on video data to enable video generation. For instance, Tune-A-Video~\cite{wu2023tune} fine-tunes a pretrained T2I diffusion model with a temporal attention mechanism in a one-shot manner. AnimateDiff~\cite{guo2023animatediff} introduces a plug-and-play motion module that integrates seamlessly into existing personalized T2I diffusion models to animate images in a similar way. However, these models rely on U-Net-based denoising networks, which have limited their performance.

Recently, some works~\cite{pixart_alpha, gao2024lumina, ma2024latte, yang2024cogvideox} have shifted the denoising network from U-Net to Diffusion Transformer, inspired by DiT~\cite{peebles2023scalable}. Video generation models with Transformers have strong spatiotemporal modeling abilities. Powered by large-scale training data, they can generate videos with rich content and motion. CogVideoX~\cite{yang2024cogvideox} utilizes a 3D Variational Autoencoder and an expert Transformer with adaptive LayerNorm to produce coherent, extended-duration videos from text prompts. However, while these methods allow text to control content, they limit the fine-grained control over object and camera movement.

\section{Method}
Our method is built on a pretrained video diffusion model~\cite{yang2024cogvideox}, consisting primarily of a diffusion transformer~\cite{peebles2023scalable} and a 3D VAE~\cite{yang2024cogvideox}. We first give a brief introduction to the process of the video diffusion model in Sec.~\ref{sec:pre}, followed by presenting the overall pipeline in Sec.~\ref{sec:pip}. In Sec.~\ref{sec:intensity}, we provide a detailed explanation of motion intensity estimation, and in Sec.~\ref{sec:injection}, we propose a new scheme for injecting motion intensity.

\subsection{Preliminaries}
\label{sec:pre}
Diffusion Transformer demonstrates superior capabilities of spatiotemporal modeling in video generation compared to U-Net architecture. In this work, we select CogVideoX~\cite{yang2024cogvideox} as the pre-trained model. Given a video $\mathbf{x} \in \mathbb{R}^{F \times H \times W \times 3}$, the 3D VAE encoder $\mathcal{E}$ compresses video frames along the spatiotemporal dimensions to obtain a latent representation $\mathbf{z_0}=\mathcal{E}(\mathbf{x})$, where $\mathbf{z_0} \in \mathbb{R}^{\left(\frac{F-1}{4} + 1\right) \times H' \times W' \times C}$.
After that, the forward diffusion and reverse denoising processes are performed in the latent space. During the forward phase, noise is incrementally added to the latent vector $\mathbf{z_0}$ over a total of $T$ steps. At each time step $t$, the diffusion process is defined as follows:
\begin{equation}
    \mathbf{z}_t = \sqrt{\overline{\alpha}_t} \mathbf{z}_0 + \sqrt{1 - \overline{\alpha}_t} \epsilon,
    \label{eq:diffusion}
\end{equation}
where $\bm{\epsilon}\in\mathcal{N}(\textbf{0},\textbf{I})$, and $\overline{\alpha}_t$ is the cumulative products of noise coefficient $\alpha_t$ at each time step. For the backward pass, a diffusion model performs iterative noise reduction, guided by the text prompt $c_{text}$ and time step $t$. The objective of this stage can be formulated as:
\begin{equation}
{L} = \expec_{\encoder(x), c_{text}, \epsilon \sim \mathcal{N}(0, 1), t }\Big[ \Vert \epsilon - \model(\mathbf{z}_t, t, c_{text}) \Vert_{2}^{2}\Big].
\label{eq:revprocess}
\end{equation}

\subsection{Overall Pipeline}
\label{sec:pip}
The framework of our model is shown in Figure~\ref{fig:framework}. The model takes a reference image, a text prompt, and two disentangled motion intensities predicted by a motion intensity estimator as inputs. During training, we first extract the first frame from the input video to use as a conditioning reference for generation. The trained motion intensity estimator then predicts the camera and object motion intensities of the input video, providing two motion scores that guide the video generation process. During inference, users can specify the desired motion intensities for the object and camera, if available, to customize the generated video. The model takes a latent $\mathbf{z}\in\mathbb{R}^{B \times T \times C \times H \times W }$ and concatenates the first frame latent of the video along the channel dimension to guide video generation. For frames beyond the first in the video sequence, zeros are padded in place. Subsequently, the model uses a text encoder to extract the features of the text prompt, which are then concatenated with the latent and fed into the diffusion transformer. Meanwhile, the two motion intensities predicted by the motion estimator are mapped to high-dimensional embeddings by MLP, then concatenated and added to the time step $t$. This combined representation serves as a modulation condition for the vision and text features, enabling fine-grained control over the motion of video generation.

\subsection{Motion Intensity Estimation}
\label{sec:intensity}
To achieve precise control over motion intensity, we train an independent motion estimator to predict the intensity of object and camera motion within a video. We provide a detailed explanation covering three aspects: the construction of training data, the design of the motion estimator architecture, and the training configuration.

\noindent\textbf{Training Data Construction.} 
Training a motion estimator to accurately predict video motion typically requires labeling object and camera motion intensities for each video—a highly challenging task. Due to the complexity of video motion, assigning specific scores to object and camera movement is impractical, as people find it difficult to consistently rate motion intensities. To address this issue, we develop a simple and intuitive labeling approach. Rather than assigning precise scores, annotators compare video pairs, indicating which video exhibits stronger object or camera motion. This method largely streamlines the annotation process. We construct 5,000 video pairs, with annotators labeling the relative motion intensities for object and camera motion within each pair.

\noindent\textbf{Motion Estimator.} 
The motion estimator needs to simultaneously predict both object motion and camera motion for video. Therefore, when designing the structure of the motion estimator, the first requirement is a backbone that can effectively represent the motion in the video. Based on this motion representation, two heads (an object motion head and a camera motion head) are introduced to map the representation to two corresponding motion intensities. Given that our video generation network is quite large, it is crucial to limit the overall parameters and computational cost of the additional motion estimator. To achieve this goal, we use TAda~\cite{huang2021tada} as the backbone for video motion representation. Given an input video $\mathbf{x}$, the motion representation of the video can be obtained through spatiotemporal modeling with TAdaConv. This process can be formulated by the following equation:
\vspace{-2mm}
\begin{equation}
\vspace{-2mm}
M = \mathrm{TAdaConv}(\mathbf{x}; \phi),
\label{eq:tada}
\end{equation}
where $M$ represents the extracted motion representation features and $\phi$ represents the parameters of TAdaConv. Afterward, we apply global average pooling over the spatiotemporal dimensions on the extracted features, followed by two separate heads: one for object motion scoring and another for camera motion scoring. Both heads are composed of MLPs. Each head predicts the respective scores for object and camera motion in the video. This process can be formulated as follows:
\vspace{-2mm}
\begin{equation}
\vspace{-2mm}
s^{\textit{object, camera}} = \text{MLP}_{\textit{object},\textit{camera}}(\text{GAP}(M); \theta),
\label{eq:object}
\end{equation}
where $s^{\textit{object}}$ and $s^{\textit{camera}}$ represent the object motion score and camera motion score, respectively, $\theta$ represents the parameters of the object or camera motion prediction head and $\text{GAP}$ denotes global average pooling.

\begin{figure*}[!t]
  \centering
    \includegraphics[width=1\linewidth]{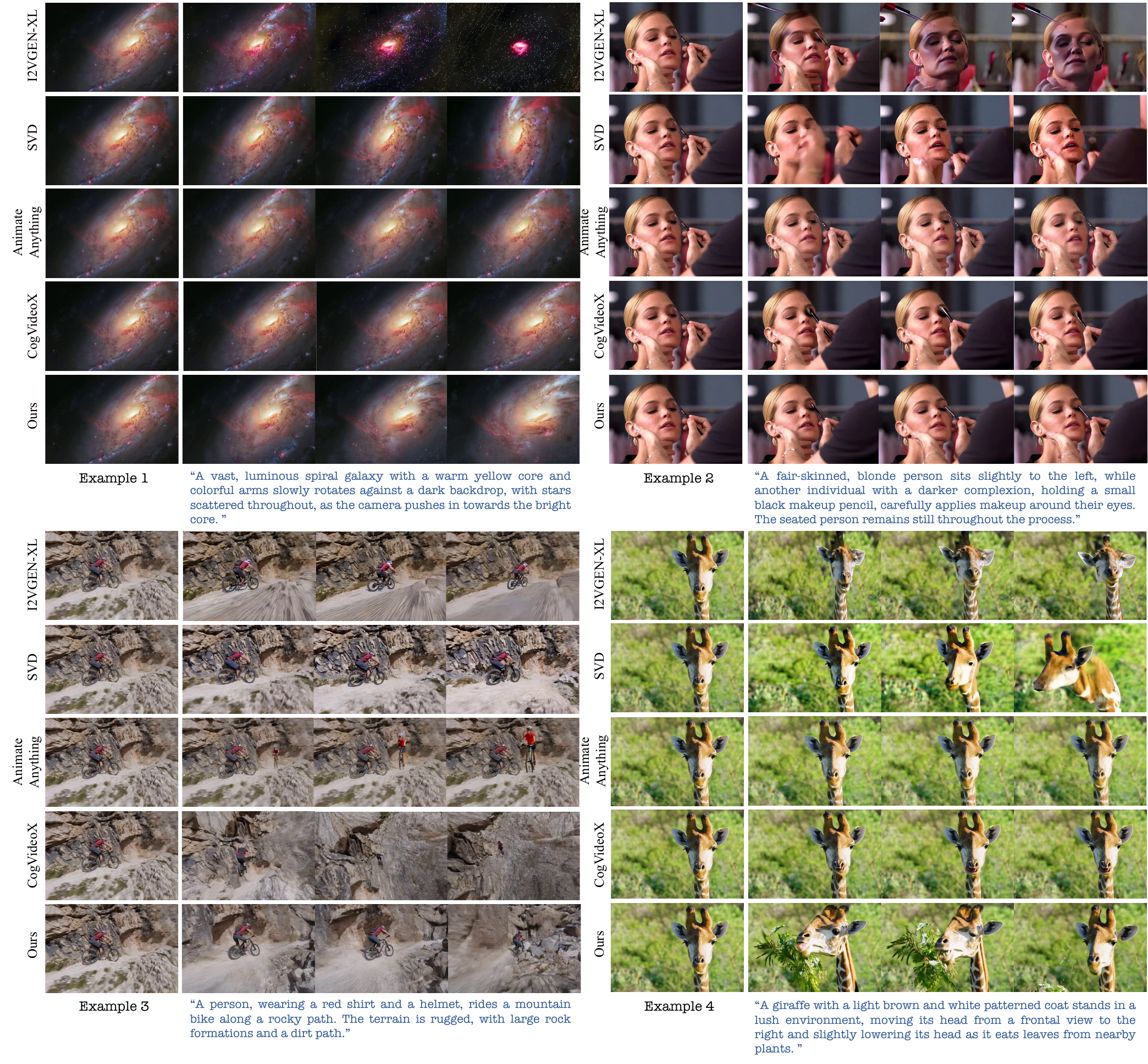}
    \vspace{-8mm}
    \caption{\textbf{Qualitative comparison with other methods.} We compare our \method with I2VGEN-XL~\cite{zhang2023i2vgen}, SVD~\cite{blattmann2023stable}, AnimateAnything~\cite{dai2023animateanything} and CogvideoX~\cite{yang2024cogvideox}. \method demonstrates superior alignment with text and image inputs compared to other methods (\textit{Example 2} and \textit{Example 4}). Additionally, as shown in 
    \textit{Example 1}, it highlights the ability of camera controlling, while other methods tend to remain static frames. \textit{Example 3} showcases the capacity of \method to control object movements, whereas other methods either remain static frames or produce unrealistic scenes that defy physical principles.}
    \label{fig:compare}
    \vspace{-3mm}
\end{figure*}

\noindent\textbf{Training Configuration.} 
Since the training dataset only contains relative motion comparison labels in the video pair, we design a contrastive learning approach to train the motion intensity estimator. This method helps the motion intensity estimator predict the relative magnitude of object and camera motion in video pairs. We train the motion estimator using the ranking loss~\cite{liu2017rankiqa}.

Specifically, given a video $\mathbf{x}$ as input to the motion estimator, we obtain object motion score $s^{\textit{object}}$ and camera motion score $s^{\textit{camera}}$ through Eq.~\ref{eq:tada} and Eq.~\ref{eq:object}. Since our goal is to learn the relative rank in the video pair, we introduce the pairwise ranking loss to train the motion estimator:
\begin{equation}
    L_{\text{o}} =\max \big(0, s^{\textit{object}}_2 - s^{\textit{object}}_1\big)
    \label{eq:rank_obj},
\vspace{-2mm}
\end{equation}
\begin{equation}
    L_{\text{c}} =\max \big(0, s^{\textit{camera}}_2 - s^{\textit{camera}}_1\big)
    \label{eq:rank_lens},
\end{equation}
here we assume that the object and camera motion of $\mathbf{x_1}$ is higher than $\mathbf{x_2}$.

However, training only with the ranking loss, the predicted scores from the motion estimator tend to cluster closely together. Such an estimator can distinguish relative motion between videos but is not practically usable as it lacks sufficient differentiation. Ideally, the predicted score should reflect clear distinctions (ranging from 1 to 10).

To make the motion estimator practically applicable, we randomly sample a subset of videos from our training dataset. Using the tracking method~\cite{xiao2024spatialtracker} combined with object masks extracted by~\cite{xie2024moving}, we calculate tracking trajectories for the object and camera motion in each video. From these trajectories, we can approximate the average motion intensity of the object $y^{\textit{object}}$ and camera $y^{\textit{camera}}$. We use them as pseudo-labels of video motion to conduct regression training for the motion estimator. The regression loss of the motion estimator training can be formulated as:
\begin{equation}
\mathcal{L}_{\text{r}} = \Vert s^{\textit{object}} - y^{\textit{object}} \Vert_{2}^{2} + \Vert s^{\textit{camera}} - y^{\textit{camera}} \Vert_{2}^{2}.
\label{eq:regression}
\end{equation}
We then jointly train the estimator using both the ranking loss and regression loss with pseudo-labels derived from the tracking results. The overall training loss can be defined:
\begin{equation}
\mathcal{L}_{\text {total }}=\mathcal{L}_{\text {o}}+ \mathcal{L}_{\text {c}} + \lambda \mathcal{L}_{\text {r}} ,
\label{eq:all}
\end{equation}
where $\lambda$ denotes the balancing parameter.
\subsection{Motion Condition Injection Design}
\label{sec:injection}
After training the motion estimator, it is crucial to inject the predicted motion intensity values into the backbone network as conditions. 
Due to the distinct meanings, these two motion types can't be directly compared and combined in the same way, so we propose a decoupled injection approach during the diffusion model training.

Specifically, we use two separate MLPs to learn high-dimensional mappings for the predicted object and camera motion vectors. These vectors are then concatenated and added to $t$, allowing two conditions to remain disentangled, and preventing ambiguity in condition injection. As shown in Figure~\ref{fig:framework}, the input motion intensities are first mapped to high-dimensional vectors, similar to $t$, and then processed through two MLPs with the same channel dimensions, respectively. The outputs are concatenated and added to $t$, collectively modulating the scaling coefficients.

\begin{table}[!t]
\setlength\tabcolsep{2pt}
\def\w{20pt} 
\caption{\textbf{Quantitative comparison} with state-of-the-art methods. We use Background Consistency to assess temporal quality, while aesthetics and imaging quality metrics are used to evaluate the visual quality of each frame.}
\vspace{-4pt}
\centering\footnotesize
\begin{tabular}{l@{\extracolsep{15pt}}c@{\extracolsep{15pt}}c@{\extracolsep{15pt}}c@{\extracolsep{15pt}}}
\shline
\multirow{2}{*}{\textbf{Method}}             & \textbf{Background}   & \textbf{Aesthetic} &    \textbf{Imaging}   \\\vspace{-4mm}\\
 & \textbf{Consistency} & \textbf{Quality} & \textbf{Quality} \\
    \shline
I2VGen-XL~\cite{zhang2023i2vgen}     & 90.93\%       & 40.14\%          & 58.35\%             \\
SVD~\cite{blattmann2023stable}      & 93.17\%       & 42.38\%          & 59.61\%             \\
AnimateAnything~\cite{dai2023animateanything}   & 93.89\%  & \underline{46.04\%}      & 61.69\%             \\
CogVideoX-5B~\cite{yang2024cogvideox}     & \underline{94.91\%}       & 45.88\%     & \underline{61.99\%}             \\
\textbf{\method}    & \textbf{95.76\%}       & \textbf{46.78\%}          & \textbf{62.29\%}             \\
\shline
\end{tabular}
\vspace{-4mm}  
  \label{tab:quantitative}%
\end{table}%

\begin{figure*}[!t]
  \centering
    \includegraphics[width=1\linewidth]{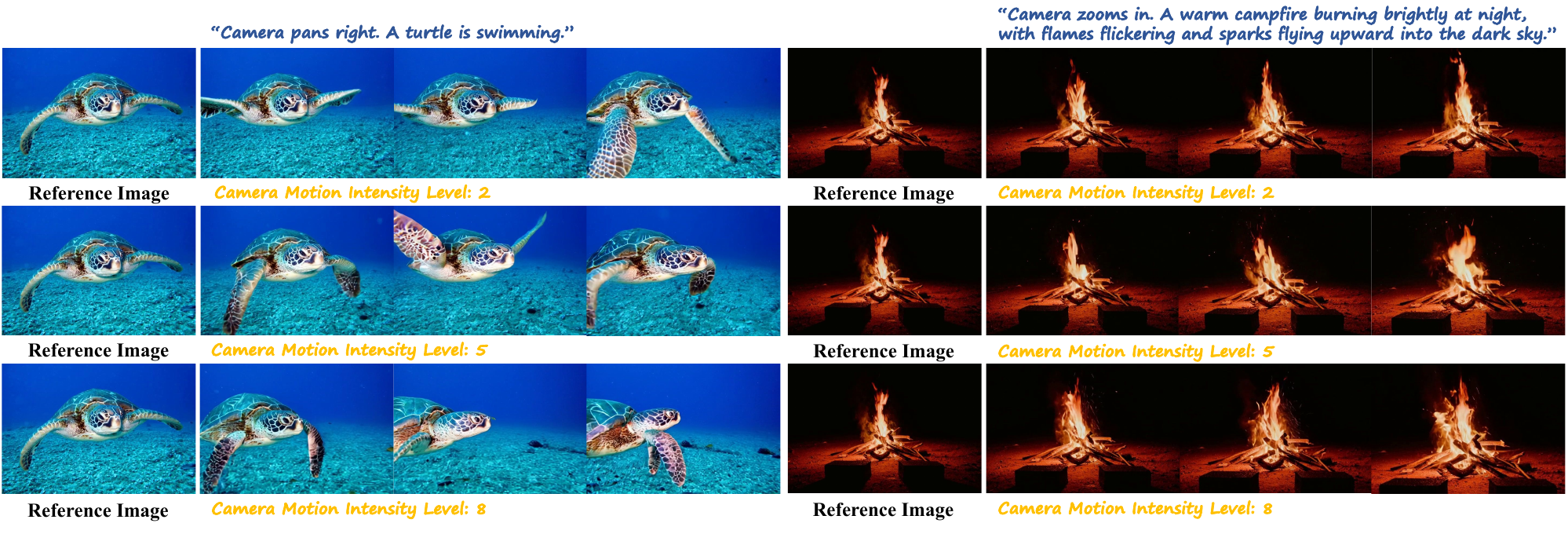}
    \vspace{-7mm}
    \caption{\textbf{Illustrations of camera motion intensity guidance.} We present two common camera movements: Zoom and Pan. Since the camera movement often impacts object motion in scenes with moving subjects, we fix the object motion intensity at $\bm{5}$ to isolate and highlight the effect of varying camera motion intensity. The camera movement becomes significant when the score increases. %
    }
    \label{fig:camera_motion_intensity}
    \vspace{-3mm} %
\end{figure*}

\begin{figure*}[!t]
  \centering
    \includegraphics[width=1\linewidth]{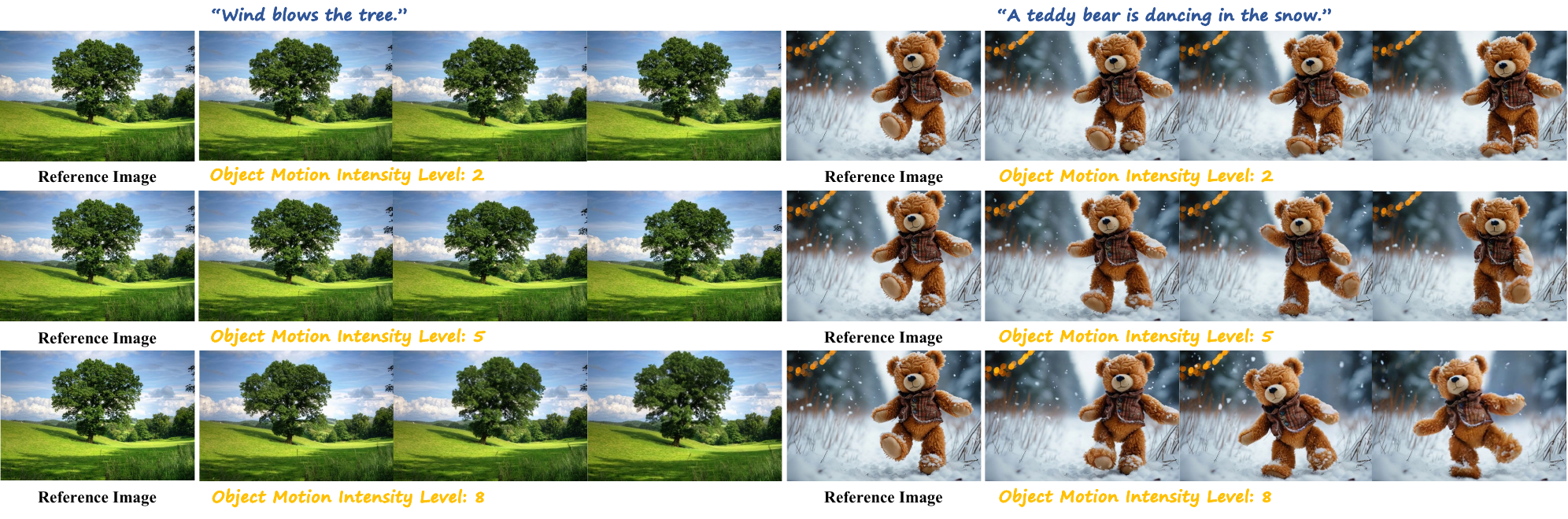}
        \vspace{-7mm}
    \caption{\textbf{Illustrations of object motion intensity guidance.} To emphasize control over object motion intensity and speed, we exclude camera motion prompts from the text and set the camera motion intensity to its minimum value of $\bm{1}$ while varying the object motion intensity. As the given object motion intensity increases, the generated video reflects a corresponding increase in object motion intensity.}
    \label{fig:object_motion_intensity}
    \vspace{-3mm}
\end{figure*}

\section{Experiments}

\subsection{Implementation Details}
\noindent\textbf{Training Configurations.} 
We implement \method using the CogVideoX~\cite{yang2024cogvideox} framework. Our model training is conducted on 100,000 high-quality videos collected by ourselves, utilizing 8 A100 GPUs with batch size 16. The training is performed using Supervised Fine-Tuning (SFT). For each training video, we sample 49 frames and apply center cropping and resizing to standardize their resolution to $480 \times 720$. We condition the Image-to-Video model training on the first frame of each video alongside its associated text prompt.

\noindent\textbf{Evaluation Metrics.} 
We conduct user studies to compare our approach with previous methods. Please refer to the appendix for detailed results. For quantitative analysis, we use the WebVID validation set~\cite{bain2021frozen}, where the first frame and corresponding prompt serve as conditions to generate videos. We employ specific metrics from VBench~\cite{huang2024vbench} to evaluate the generated videos, using Background Consistency to assess temporal quality, while aesthetics and imaging quality metrics are used to evaluate the visual quality of each frame.

\subsection{Comparisons with Existing Alternatives}
We compare \method with several recent Image-to-Video (I2V) methods. I2VGEN-XL~\cite{zhang2023i2vgen} and AnimateAnything~\cite{dai2023animateanything} are classic I2V approaches that enable video generation conditioned on a given image and text. AnimateAnything also supports coarse control over motion intensity. SVD~\cite{blattmann2023stable} is a widely used I2V model that employs U-Net as its denoising network. Additionally, CogVideoX~\cite{yang2024cogvideox} is an open-source video generation model based on the Diffusion Transformer.

\noindent\textbf{Quantitative Results.} 
We conduct quantitative experiments using WebVID~\cite{webvision} validation dataset. Specific metrics from VBench~\cite{huang2024vbench} are selected to evaluate the experimental results. Among these, the Background Consistency metric from the CLIP Score~\cite{radford2021learning} effectively reflects the generative quality in the temporal dimension of video. Meanwhile, Aesthetic Quality and Imaging Quality respectively assess the aesthetic appeal and the quality of each individual frame in the generated video. It is important to note that we do not utilize the prompt suite from VBench; instead, we only employ their evaluation procedures and models. 
As shown in~\tablename~\ref{tab:quantitative}, compared to U-Net-based generative models like SVD, I2VGEN-XL, and AnimateAnything, our approach demonstrates significant improvements in both temporal consistency and visual quality of the generated videos. Even relative to our baseline, CogVideoX, our method achieves noticeable enhancements. This is largely due to our model’s integration of motion intensity prediction and decoupled conditional injection, which effectively reduces the ambiguity between the motion described in text prompts and the actual motion intensity in the generated videos. This also demonstrates that precise motion intensity control signals can help video models converge more effectively.

\begin{table}[!t]
\setlength\tabcolsep{2pt}
\def\w{20pt} 
\caption{\textbf{Ablation Study for proposed modules.} Motion Estimator (M), Decoupled injection strategy (D). SSIM and S mean previous motion modeling methods: inter-frame SSIM~\cite{chen2025livephoto} and feature difference~\cite{dai2023animateanything} respectively.}
\vspace{-4pt}
\centering\footnotesize
\begin{tabular}{l@{\extracolsep{15pt}}c@{\extracolsep{15pt}}c@{\extracolsep{15pt}}c@{\extracolsep{15pt}}}
\shline
\multirow{2}{*}{\textbf{Method}}             & \textbf{Background}   & \textbf{Aesthetic} &    \textbf{Imaging}   \\\vspace{-4mm}\\
 & \textbf{Consistency} & \textbf{Quality} & \textbf{Quality} \\
    \shline
\method \textit{w/o} M     & \underline{95.13\%}       & 45.61\%          & 60.15\%             \\
\method \textit{w/} S    & 94.97\%       & 46.13\%          & \underline{60.73\%}             \\
\method \textit{w/} SSIM    & 92.99\%       & 45.72\%          & 54.75\%             \\
\method \textit{w/o} D      & 94.03\%       & \underline{46.27\%}          & 58.73\%             \\
\textbf{\method}     & \textbf{95.76\%}       & \textbf{46.78\%}          & \textbf{62.29\%}  \\
\shline
\end{tabular}
  \label{tab:ablation}%
\end{table}%

\begin{table}[!t]
\setlength\tabcolsep{20pt}
\def\w{20pt} 
\caption{\textbf{Comparison with previous motion intensity estimation methods.} We calculate and compare object and camera motion scores for each video pair, then validate these predictions against manually annotated ground truth. Correct predictions score 1 point. Evaluation is conducted on the validation set of the video pair dataset introduced in Sec.~\ref{sec:intensity}. }
\vspace{-4pt}
\centering\footnotesize
\begin{tabular}{c@{\extracolsep{20pt}}c}
\shline
\textbf{Method} & \textbf{Motion Estimation Accuracy} \\
\midrule
SSIM     & 44.56\%       \\
Ours    & 72.80\%       \\
\shline
\end{tabular}
\vspace{-4mm}
\label{tab:estimator}%
\end{table}

\noindent\textbf{Qualitative Analysis.} 
In Figure~\ref{fig:compare}, we select a set of representative samples to qualitatively compare \method with I2VGEN-XL~\cite{zhang2023i2vgen}, SVD~\cite{blattmann2023stable}, AnimateAnything~\cite{dai2023animateanything}, and CogVideoX~\cite{yang2024cogvideox}. We select cases involving people, animals, natural scenes, and fast-moving scenarios. As can be seen, the identity of the subject in the videos generated by I2VGEN-XL is not well-preserved, and there are occasional discrepancies between the motion and the text prompt. SVD also appears to face issues with preserving the identity of objects, and in fast-moving scenarios, the generated video exhibits limited motion intensity (e.g., the bicycle remains stationary). Although the video frames generated by AnimateAnything are well-aligned with the input image, in most scenes, the generated video is almost static, and there are occasional interruptions from other objects that interfere with the main subject. Compared to previous methods, CogVideoX shows some improvements in motion continuity. However, it occasionally fails to align with the content of the input image and exhibits limited motion. In Example 3, it generates content that does not adhere to the physical rules of the real world.

In contrast, \method is capable of generating videos that align well with both the input image and text. The generated videos exhibit substantial camera and object motion, producing visually appealing shots and motion that adhere to the laws of the physical world.

\subsection{Ablation Studies}
In this section, we provide detailed analyses of proposed modules. We begin with quantitative experiments on all of the proposed modules, followed by a qualitative analysis of the controllability of object and camera motion intensities. 
Finally, we evaluate the motion magnitude between video pairs in the validation set. Specifically, we compare the accuracy of our motion estimator against SSIM to verify its predictive effectiveness and its ability to decouple motion.

\noindent\textbf{Motion Estimator.} 
To validate the overall effectiveness of the proposed motion estimator, we compare the quantitative performance of \method trained with fixed motion intensity (set to a default value of 5) versus \method trained with motion intensities estimated by the motion estimator. The experiments are conducted on the WebVID validation set. As shown in~\tablename~\ref{tab:ablation}, the quality of videos generated by \method without the motion estimator (\method \textit{w/o} M) shows a noticeable decline. This is due to the variability in object and camera motion within the training data, using a fixed intensity value confuses the model's understanding of video motion dynamics.

\noindent\textbf{Decoupled Motion Condition Injection.} 
To quantitatively demonstrate the effectiveness of the proposed decoupled injection method, we compare the performance of \method trained with decoupled versus non-decoupled motion intensity injection. In the non-decoupled injection approach, object and camera motion are not specifically separated along feature channels but are instead mixed and injected together.
As shown in~\tablename~\ref{tab:ablation}, the performance of \method with non-decoupled injection (\method \textit{w/o} D) is inferior to that of the decoupled injection approach. This is primarily because object and camera motion occur in different spatial dimensions; mixing them together obscures their distinct contributions, making it challenging for the model to discern each type of motion, thus complicating the training process.

\noindent\textbf{Comparison with Previous Motion Intensity Estimation Methods.} 
We further compare our method with previous methods for modeling motion intensity, which uses inter-frame SSIM~\cite{chen2025livephoto} (\method \textit{w/} SSIM) or feature difference~\cite{dai2023animateanything} (\method \textit{w/} S). Models are trained following these methods; however, since neither approach can decouple object and camera motion, we use a single motion intensity guidance to train the model. As shown in Table~\ref{tab:ablation}, both methods exhibit varying degrees of performance decline. This is because neither SSIM nor inter-frame feature difference aligns well with human perception of video motion intensity. Additionally, both methods fail to decouple complex video motion dynamics, instead modeling the motion intensity of the entire scene as a coarse, unified value, which leads to inaccurate estimations.

We further evaluate the trained motion estimator and the SSIM-based motion intensity estimation method on the validation set of the manually constructed video pair dataset introduced in Sec.~\ref{sec:intensity}. Since SSIM cannot decouple object and camera motion, we use its predicted overall motion intensity as a proxy for both object and camera motion intensities. We calculate the object and camera motion scores for both videos in a video pair and compare their relative magnitudes. The obtained comparison is then compared with the manually annotated ground truth (GT). If the predicted object or camera motion relationship is correct, it is scored as 1 point. The final motion intensity relationship prediction accuracy is calculated using this method, as shown in~\tablename~\ref{tab:estimator}. Our motion estimator achieves excellent accuracy in predicting motion relationships, surpassing SSIM-based method by 28\%. This demonstrates that the trained motion estimator effectively decouples object and camera motion in videos.

\noindent\textbf{Motion Intensity Estimation.} 
As demonstrated in Sec.~\ref{sec:intensity}, we represent the intensity of object and camera motion in a video as a score, reflecting the magnitude and speed of both object and camera movements in the video. We perform ablation analysis of camera and object motion intensities in Figure~\ref{fig:camera_motion_intensity} and Figure~\ref{fig:object_motion_intensity}, respectively.
In Figure~\ref{fig:camera_motion_intensity}, we control for camera motion and explore the impact of intensity control on two representative types of camera movements: zoom and pan. Since camera movement often influences object motion in scenes with moving subjects, we fix the object motion intensity at 5 to highlight the effect of controlling camera movement by varying its intensity level.
From the left set of images, we observe that when the camera motion intensity level is set to 2, the rightward pan of the camera is limited. As the motion intensity level increases, the amplitude of the rightward pan significantly grows, resulting in a more substantial shift in perspective.
For the images on the right, as the camera motion intensity level increases, the degree of camera zoom-in also increases. Figure~\ref{fig:camera_motion_intensity} demonstrates that \method while maintaining the intensity and speed of object motion, is capable of customizing the intensity and speed of camera motion.
In Figure~\ref{fig:object_motion_intensity}, to highlight the control over object motion intensity and speed, we do not include camera motion prompts in the text and set the camera motion intensity level to the minimum value of 1, while varying the object motion intensity level. From the two sets of images, it is clear that as the given object motion intensity level increases, the speed and intensity of object motion in the generated video both become faster and larger.

\section{Conclusion}
In this work, we propose \method, a general image-to-video (I2V) generation framework that enables decoupled modeling and control of video motion. 
To achieve this, we train a dedicated motion estimator that directly predicts object and camera motion intensities in line with human perception. To address the challenge that human annotators cannot directly label absolute motion intensities, we develop a novel annotation method for video pairs specifically for training the motion estimator. We design a motion estimator with a backbone for video motion representation and disentangled heads to predict object and camera motion, trained using a contrastive learning strategy. Finally, we inject the predicted motion intensities into a diffusion model, thereby improving training convergence and user customization ability. This entire pipeline demonstrates impressive performance across diverse domains and task instructions.

\nopagebreak
{
    \small
    \bibliographystyle{ieeenat_fullname}
    \bibliography{main}
}
\renewcommand{\thefootnote}{\arabic{footnote}}

\clearpage
\setcounter{page}{1}
\maketitlesupplementary
\appendix

\section{Implementation Details}
We supplement more details of the training of motion estimator. For training the motion estimator, we utilize 8 A100 GPUs with batch size 64. The learning rate is set to $5\times10^{-6}$. To align with the training configuration of \method, input videos are cropped to a resolution of $480\times720$ and sampled to 49 frames. The motion estimator is trained for 10,000 steps using the Adam optimizer with $\beta_1 = 0.9$ and $\beta_2 = 0.999$. We set the weight of regression loss $\lambda$ to 0.1.
\section{Details on the Training Data for Motion Estimator}
\label{sec:motiondata}
In this section, we provide more details on the training data for the motion estimator. We ask 15 annotators to participate in this annotation process. The annotators are asked to label video pairs from several aspects: First, they are asked to determine whether the two videos in a pair contain a moving object. A video is considered to have a moving object only if it features a foreground object in motion. Meanwhile, camera motion focuses on the global motion in the scene. If a video in the pair contains a moving object, it is labeled as 1; otherwise, it is labeled as 0. Note that comparisons of object motion between the two videos are only made when at least one video in the pair features a moving object. Next, annotators are tasked with labeling the relative magnitude of the object and camera motion in each video pair. If both videos contain object or camera motion, the corresponding item is annotated based on the annotators’ subjective judgment. If only one video in the pair exhibits object or camera motion, the video with motion is considered significantly greater in the respective category. Specifically, we define the annotations as follows: if the first video shows significantly greater camera or object motion than the second one, it is labeled as 2; if it is only slightly greater, it is labeled as 1. Conversely, if the first video shows significantly or slightly less motion, it is labeled as -2 or -1, respectively. If neither video exhibits object or camera motion, the corresponding item is labeled as 0. During the training process using contrastive learning, this label is employed to amplify the motion differences between two videos. If a specific motion in the first video is significantly greater than that in the second, the corresponding loss is set to twice that of cases with a smaller difference.

After completing one round of annotation, we conduct a sampling check on 5,000 video pairs, reviewing $20\%$ of them. The investigation achieves an accuracy rate of $95\%$, meeting the annotation standards. This demonstrates that the annotated data aligns well with human perception of the relative magnitude of object and camera motion in videos.

\section{User Study on Comparisons with Existing Alternatives}
\label{sec:user}
Since the metrics in VBench~\cite{huang2024vbench} cannot fully evaluate the performance of the model, we conduct user studies. We ask 10 annotators to participate in this process. To ensure the generalization of the evaluation, we select a wide variety of real and animated images, including elements such as people, animals, camera movement, plants, and natural landscapes. Twenty image-text prompts are selected and processed by each compared method, including \method, generating a total of 100 video clips. Each participant is presented with two videos generated by different methods for the same prompts and asked to choose the one that performed better in four aspects: \textit{Text Consistency} evaluates if the motion and content follow the text prompt. \textit{Image Consistency} assesses the ability to preserve the identity of the reference image. \textit{Content Quality} determines the overall quality of video generation, including visual appeal, definition, and the logical coherence of the generated content. \textit{Motion Quality} evaluates the plausibility and richness of the motion. The pairwise comparison is repeated for all combinations of videos, resulting in $C^5_2$ comparisons. 

As shown in Tab.~\ref{tab:user_study}, our method demonstrates superior performance, particularly in terms of Text Consistency, Content Quality and Motion Quality. This highlights the effectiveness of our approach in text-based motion control and the generation of videos with content and motion that align more closely with human perception.
\begin{table*}[!t]
\setlength\tabcolsep{2pt}
\def\w{20pt} 
\caption{%
    \textbf{Results of user study.} The best results for each column are \textbf{bold}. We ask annotators to rate videos based on four aspects: Text Consistency, which assesses how well the motion and content adhere to the textual descriptions; Image Consistency, which evaluates the ability to preserve the identity of the reference image; Content Quality, which focuses on inter-frame coherence and definition; and Motion Quality, which measures the plausibility and richness of the motion.
}
\vspace{-8pt}
\centering\footnotesize
\begin{tabular}{l@{\extracolsep{10pt}}c@{\extracolsep{10pt}}c@{\extracolsep{10pt}}c@{\extracolsep{10pt}}c@{\extracolsep{10pt}}c}
\shline
Method        & I2VGEN-XL     & SVD & AnimateAnything & CogVideoX-5B      & \textbf{\method} \\
\shline
Text Consistency $\uparrow$   & 32.50\% & 39.38\%    &   25.00\%     & 63.13\% & \textbf{90\%} \\
Image Consistency  $\uparrow$  &  27.50\% &  36.88\%  &  56.25\%   &  62.50\% & \textbf{66.88\%}\\
Content Quality $\uparrow$  &  31.25\%  & 45.63\% &  33.13\%   &  63.13\%  & \textbf{76.88\%} \\

Motion Quality $\uparrow$  &  26.25\% & 48.13\%  & 39.38\%  &  61.25\%  & \textbf{75.00\%}     \\

\shline
\end{tabular}
  \label{tab:user_study}%
\end{table*}%

\begin{figure*}[!t]
  \centering
    \includegraphics[width=1\linewidth]{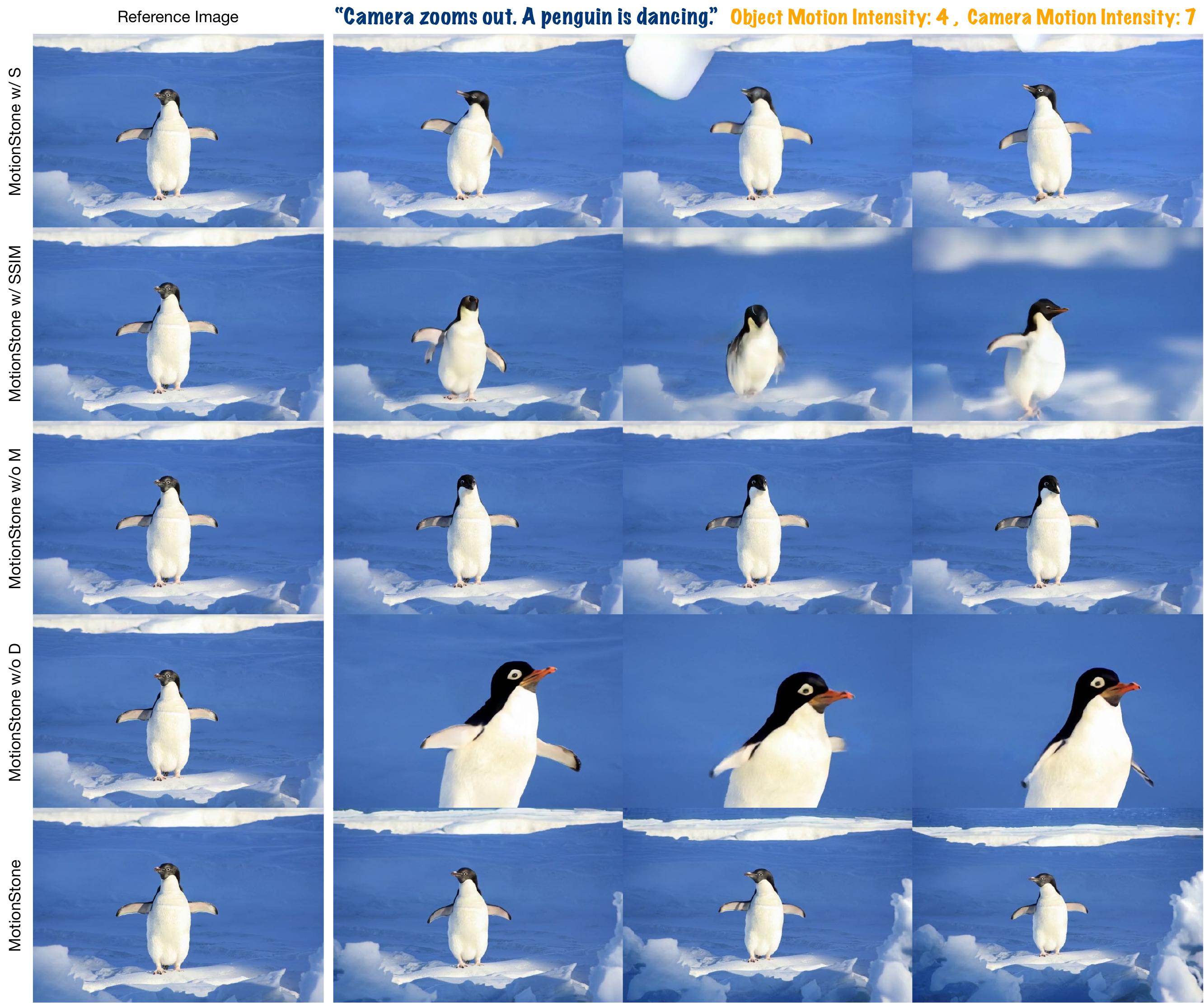}
        \vspace{-7mm}
    \caption{\textbf{Qualitative ablation for proposed modules.} Using inter-frame SSIM~\cite{chen2025livephoto} and feature difference~\cite{dai2023animateanything} (\method \textit{w/} SSIM and \method \textit{w/} S) causes varying degrees of unnatural background motion (In the first row, the snow block in the upper left corner of the third column appears. In the second row, background motion blur is observed.) and does not follow the camera motion described in the text prompt. Omitting the proposed motion estimator (\method \textit{w/o} M) and the decoupled injection method (\method \textit{w/o} D) results in issues such as generating static video and confusion or overlap between camera motion and object motion control, respectively. These approaches also fail to follow the camera motion described in the text prompt successfully.}
    \label{fig:abla_qualititive}
    \vspace{-3mm}
\end{figure*}

\section{Evaluation Metrics}
We select several metrics from VBench~\cite{huang2024vbench} for quantitative evaluation experiments, including \textit{Background Consistency}, \textit{Aesthetic Quality}, \textit{Imaging Quality}, \textit{Subject Consistency}, \textit{Motion Smoothness}, \textit{Dynamic Degree} and \textit{Temporal Flickering}. It is important to note that, we utilize only its models and evaluation processes, excluding its prompt suite. Consequently, some metrics that strictly require the use of the prompt suite are omitted. The detailed information on each metric is introduced as follows.

\noindent\textbf{Background Consistency.} 
This metric measures the temporal consistency of the background scenes by calculating CLIP~\cite{radford2021learning} feature similarity across frames.

\noindent\textbf{Aesthetic Quality.}
This metric assesses the human-perceived artistic and aesthetic value of each video frame utilizing the LAION aesthetic predictor. This tool captures various aesthetic dimensions, including composition, color richness and harmony, photorealism, naturalness, and the artistic quality of the video frames.

\begin{table*}[!t]
\setlength\tabcolsep{2pt}
\def\w{20pt} 
\caption{%
    \textbf{More quantitative ablation results on VBench~\cite{huang2024vbench}.} The best results for each column are \textbf{bold}. Motion Estimator (M), Decoupled injection strategy (D). SSIM and S mean previous motion modeling methods: inter-frame SSIM~\cite{chen2025livephoto} and feature difference~\cite{dai2023animateanything} respectively.
}
\vspace{-8pt}
\centering\footnotesize
\begin{tabular}{l@{\extracolsep{10pt}}c@{\extracolsep{10pt}}c@{\extracolsep{10pt}}c@{\extracolsep{10pt}}c@{\extracolsep{10pt}}c@{\extracolsep{10pt}}c@{\extracolsep{10pt}}c@{\extracolsep{10pt}}c}
\shline
\multirow{2}{*}{\textbf{Method}}             & \textbf{Background}   & \textbf{Aesthetic} &    \textbf{Imaging}  & \textbf{Subject} & \textbf{Motion} & \textbf{Dynamic} & \textbf{Temporal}  \\\vspace{-4mm}\\
 & \textbf{Consistency} & \textbf{Quality} & \textbf{Quality} & \textbf{Consistency}& \textbf{Smoothness}& \textbf{Degree}  & \textbf{Flickering} \\
    \shline
\method \textit{w/o} M     & 95.13\%      & 45.61\%          & 60.15\%          & 93.34\%                & 98.51\%  & 43\%          & 96.51\%                      \\
\method \textit{w/o} S     & 94.97\%       & 46.13\%          & 60.73\%          & 92.99\%       & 98.48\%          & 42\%          & 96.42\%                   \\
\method \textit{w/} SSIM   & 92.99\%       & 45.72\%          & 54.75\%    & 88.96\%                & 97.51\%          & 47\%          & 93.54\%                  \\
\method \textit{w/o} D        & 94.03\%       & 46.27\%          & 58.73\%          & 92.54\%                & 97.59\%          & 48\%   & 95.20\%                     \\
\hline
\textbf{\method}     & \textbf{95.76\%}       & \textbf{46.78\%}          & \textbf{62.29\%} & { \textbf{94.56\%}} & \textbf{98.96\%} & \textbf{48\%} & \textbf{97.41\%}  \\
\shline
\end{tabular}
  \label{tab:abla_quantitative}%
\end{table*}%

\begin{figure*}[!t]
  \centering
    \includegraphics[width=1\linewidth]{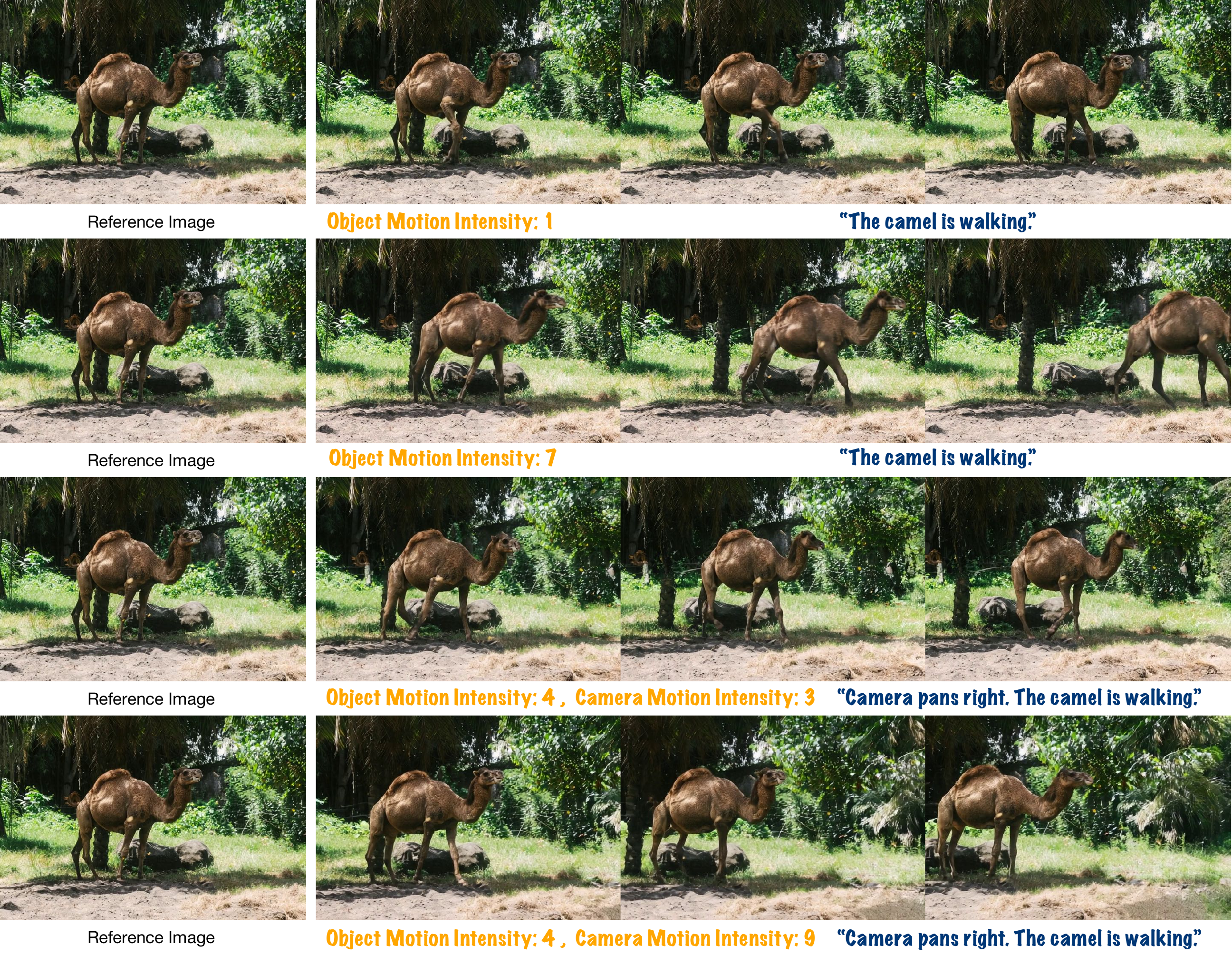}
        \vspace{-7mm}
    \caption{\textbf{Illustrations of object and camera motion intensity guidance.} \method can decouple and independently control camera motion and object motion intensities. When either camera motion or object motion is increased, the generated videos exhibit excellent adherence to the respective motion changes.}
    \label{fig:motion_intensity}
    \vspace{-3mm}
\end{figure*}

\noindent\textbf{Imaging Quality.}
Imaging quality pertains to distortions such as over-exposure, noise, and blur observed in the generated frames. This metric measures this using the MUSIQ~\cite{ke2021musiq} image quality predictor, which is trained on the SPAQ~\cite{fang2020perceptual} dataset.

\noindent\textbf{Subject Consistency.}
This metric calculates the DINO~\cite{caron2021emerging} feature similarity across frames to evaluate the consistency of a subject's appearance throughout the video.

\noindent\textbf{Motion Smoothness.}
Evaluating the smoothness of motion in generated videos and its adherence to real-world physical laws is crucial. To assess this, this metric leverages motion priors from the video frame interpolation model~\cite{li2023amt}.

\noindent\textbf{Dynamic Degree.}
As a completely static video might perform well in the previously mentioned temporal quality metrics, it is essential to assess the level of dynamics (i.e., the presence of significant motions) in the generated videos. To achieve this, this metric uses RAFT~\cite{teed2020raft} to estimate the extent of dynamics in the synthesized outputs.

\noindent\textbf{Temporal Flickering.}
Generated videos may display imperfect temporal consistency, particularly in local and high-frequency details. To quantify this, this metric extracts static frames and calculates the mean absolute difference between them.

\section{Limitation}
\label{sec:limitation}

Although \method has made notable progress in I2V generation and motion intensity control, it still faces several limitations. 
First, \method is built upon CogVideoX, and due to constraints in memory and computational resources, it can only generate videos of approximately 6 seconds in length at a specific resolution. We believe that as the computational demands of foundational video generation models decrease in the future, \method will be able to generate longer videos with higher resolutions. Furthermore, with reduced computational resource requirements, it will be feasible to design a larger motion estimator and leverage more extensive training datasets to develop a more powerful model. The enhanced motion estimator could better assist I2V generation, and we are confident that such advancements will lead to superior performance.

\begin{table}[!t]
\setlength\tabcolsep{20pt}
\def\w{20pt} 
\caption{\textbf{Ablation on motion intensity guidance.} Compared to previous methods, our motion estimator achieves more precise control over motion intensity, generating videos with camera or object motion that better aligns with user requirements.}
\vspace{-4pt}
\centering\footnotesize
\begin{tabular}{c@{\extracolsep{20pt}}c}
\shline
\textbf{Method} & \textbf{Motion Strength Error} \\
\midrule
Feature Difference (S)~\cite{dai2023animateanything}  & 11.55 \\
SSIM~\cite{chen2025livephoto}     & 11.27       \\
\textbf{Ours}    & \textbf{2.52}      \\
\shline
\end{tabular}
\vspace{-4mm}
\label{tab:motion_strength}%
\end{table}
\section{More Experiments}
In this section, we first present additional ablation studies, including more detailed qualitative and quantitative experiments, as well as an evaluation of the motion strength error of our proposed motion estimator compared to previous motion intensity estimation methods. Subsequently, we provide more specific quantitative comparison results. Finally, we provide additional cases to showcase the generative capabilities of \method.

\begin{figure*}[!t]
  \centering
    \includegraphics[width=1\linewidth]{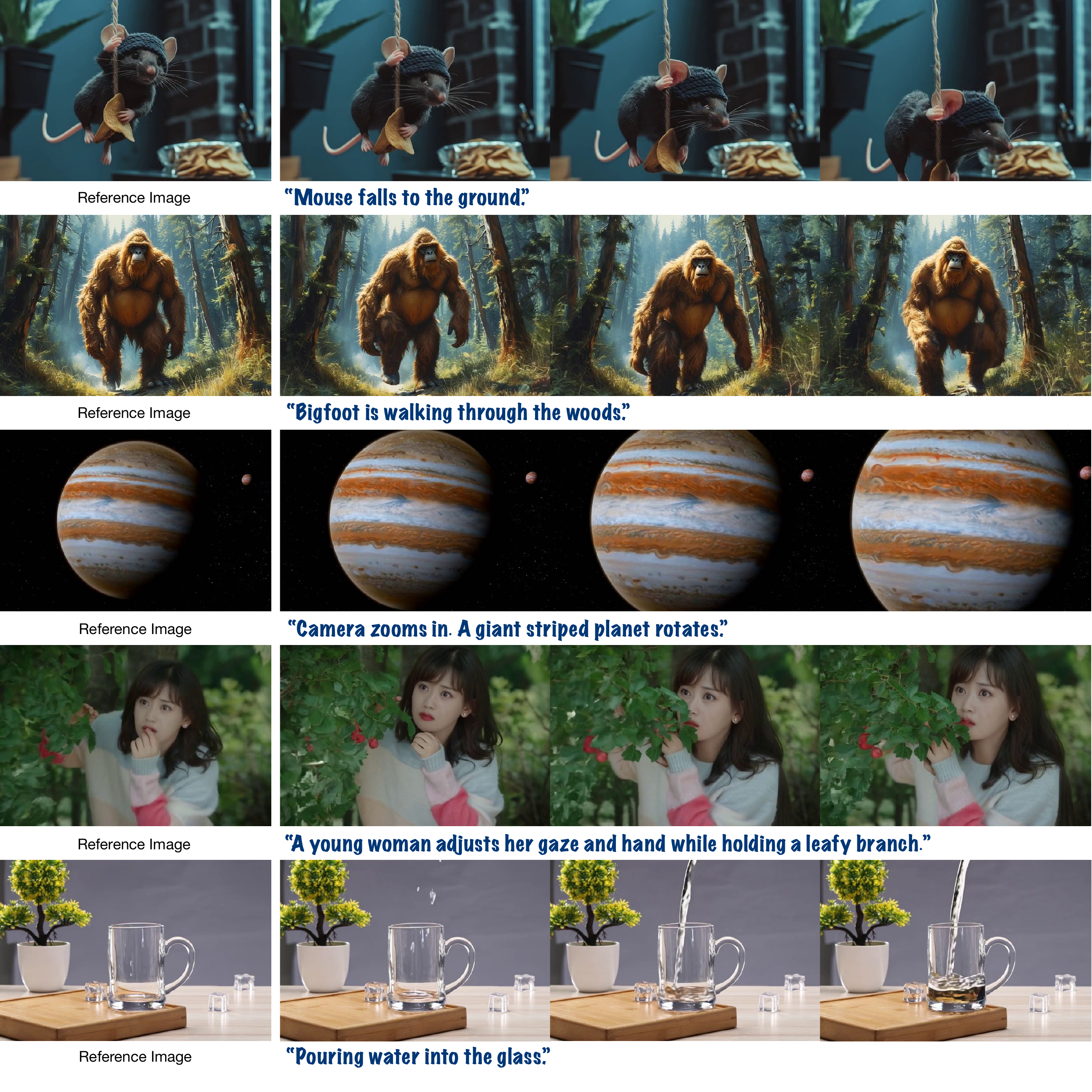}
        \vspace{-7mm}
    \caption{\textbf{More cases generated by \method.} \method demonstrates impressive generation quality across various scenarios. Here, the default object motion intensity or camera motion intensity (if applicable) is set to 5.}
    \label{fig:case}
    \vspace{-3mm}
\end{figure*}

\begin{table*}[!t]
\setlength\tabcolsep{2pt}
\def\w{20pt} 
\caption{%
    \textbf{More quantitative comparison results on VBench~\cite{huang2024vbench}}. The best results for each column are \textbf{bold}.
}
\vspace{-8pt}
\centering\footnotesize
\begin{tabular}{l@{\extracolsep{10pt}}c@{\extracolsep{10pt}}c@{\extracolsep{10pt}}c@{\extracolsep{10pt}}c@{\extracolsep{10pt}}c@{\extracolsep{10pt}}c@{\extracolsep{10pt}}c@{\extracolsep{10pt}}c}
\shline
\multirow{2}{*}{\textbf{Method}}             & \textbf{Background}   & \textbf{Aesthetic} &    \textbf{Imaging}  & \textbf{Subject} & \textbf{Motion} & \textbf{Dynamic} & \textbf{Temporal}  \\\vspace{-4mm}\\
 & \textbf{Consistency} & \textbf{Quality} & \textbf{Quality} & \textbf{Consistency}& \textbf{Smoothness}& \textbf{Degree}  & \textbf{Flickering} \\
    \shline
I2VGen-XL~\cite{zhang2023i2vgen}     & 90.93\%          & 40.14\%                & 58.35\%          & 86.97\%                & 97.02\%  & 44\%          & 95.24\%                      \\
SVD~\cite{blattmann2023stable}     & 93.17\%          & 42.38\%                & 59.61\%          & 93.23\%       & 97.39\%          & 40\%          & 94.70\%                   \\
AnimateAnything~\cite{dai2023animateanything}  & 93.89\%          & 46.04\%      & 61.69\%    & 93.72\%                & 97.58\%          & 4\%          & 95.48\%                  \\
CogVideoX-5B~\cite{yang2024cogvideox}       & 94.91\%          & 45.88\%                & 61.99\%          & 94.39\%                & 98.76\%          & 36\%   & 96.73\%                     \\
\hline
\textbf{\method}     & \textbf{95.76\%} & { \textbf{46.78\%}} & \textbf{62.29\%} & { \textbf{94.56\%}} & \textbf{98.96\%} & \textbf{48\%} & \textbf{97.41\%}  \\
\shline
\end{tabular}
  \label{tab:more_evaluation_results}%
\end{table*}%
\subsection{More Ablations}
\noindent\textbf{More Quantitative and Qualitative Results.}
We first supplement additional quantitative metrics on VBench~\cite{huang2024vbench} to demonstrate the superiority of \method. As shown in Tab.~\ref{tab:abla_quantitative}, benefiting from the support of the motion estimator and the decoupled injection method, \method outperforms other motion intensity modulation approaches and models without these strategies in terms of generated quality, inter-frame consistency of subjects and backgrounds, motion magnitude, and temporal quality.

Furthermore, we conduct qualitative ablation studies. As shown in Fig.~\ref{fig:abla_qualititive}, we generate videos using prompts containing both camera and object motions. We observe that \method \textit{w/} S and \method \textit{w/} SSIM fail to follow the camera motion described in the text prompt. Additionally, \method \textit{w/} S exhibits unnatural motion in background objects (e.g., the snow block in the upper left corner of the third column), while \method \textit{w/} SSIM displays motion blur issues. These problems are common to non-decoupled motion intensity modulation methods, as they inadvertently cause undesirable background motion while animating the subject.
We observe that the \method \textit{w/o} M model, which does not utilize the motion estimator, generates static frames without responding to the specified motion intensity. This issue arises because, during training, the model does not receive varying signals corresponding to different motion intensities but rather a constant signal. As a result, the model fails to interpret the provided intensity control signals and is unable to model motion intensity accordingly.
\method \textit{w/o} D exhibits excessive motion, affecting both the object and the camera motion. Moreover, it fails to follow the text prompt to perform a zoom-out motion, instead generating an opposite camera motion. This issue stems from the lack of decoupled injection of camera and object motion intensity signals. Without clear separation, the model struggles to associate the signals with the specific motion components they are meant to control, leading to unpredictable overlap or confusion. Consequently, the generated video lacks coherent and orderly control. In contrast, \method accurately follows the object and camera motion descriptions provided in the text prompt and generates visually appealing and motion-consistent videos based on the specified motion intensities. This demonstrates the effectiveness of the proposed modules.

\noindent\textbf{Motion Intensity Guidance.}
We provide an additional example to demonstrate the decoupled control capabilities of \method for object motion and camera motion intensities. As shown in Fig.~\ref{fig:motion_intensity}, in the first two rows, the text prompt does not specify camera motion, so the camera motion intensity is set to the minimum. By increasing the control of object motion intensity, it is evident that the camel moves faster. In contrast, in the last two rows, we introduce camera motion descriptions in the text prompt and adjust the camera motion intensity while reducing the control of object motion intensity. It is observable that as the object motion intensity decreases from 7 to 4, the camel slows down. Meanwhile, as the camera motion intensity increases from 3 to 9, the camera pans to the right more quickly. These examples strongly demonstrate the ability of the \method to decouple and independently control camera and object motions in generated videos.

Furthermore, we compare the performance of different motion intensity guidance methods. Using predefined motion intensity values, we generate videos and subsequently apply a motion estimator to obtain the corresponding motion intensities. The mean squared error (MSE) between the generated video intensities and the input values is then calculated. As shown in Tab.~\ref{tab:motion_strength}, the motion estimator proposed in this work provides more stable motion guidance and ensures that the motion intensities in the generated videos align more closely with the user-specified values.

\subsection{More Results}
We supplement additional quantitative comparison results across more evaluation dimensions on VBench~\cite{huang2024vbench}, as shown in Tab.~\ref{tab:more_evaluation_results}. \method demonstrates superior performance in terms of temporal quality and motion magnitude of the generated videos compared to previous methods.

We also provide additional examples generated by \method, as shown in Fig.~\ref{fig:case}. These include real human figures, anime-style characters, animals, and natural scenes. \method demonstrates remarkable capabilities in conjuring entirely new content out of thin air.

We provide the original video cases showcased in the paper within the supplementary materials. The detailed video effects can be found in the designated folder.

\end{document}